\documentclass[runningheads]{llncs}
\usepackage[T1]{fontenc}

\usepackage[capitalize,noabbrev]{cleveref}
\usepackage{url}
\usepackage{todonotes} 
\usepackage{graphicx}
\usepackage[normalem]{ulem} 
\usepackage{xcolor}

\begin{document}
\setlength{\tabcolsep}{4pt} 

\title{
ParCzech4Speech: A New Speech Corpus Derived from Czech Parliamentary Data%
\thanks{This preprint has not undergone peer review or any post-submission improvements or corrections. The Version of Record of this contribution is published in Proceedings of the 28th International Conference on Text, Speech, and Dialogue (TSD 2025), and is available online at https://doi.org/10.1007/978-3-032-02548-7\_25.}
}
\titlerunning{ParCzech4Speech}
%
\author{Vladislav Stankov\inst{1}\orcidID{0000-0002-2034-6485} \and
Matyáš Kopp\orcidID{0000-0001-7953-8783} \and
Ondřej Bojar\orcidID{0000-0002-0606-0050}
}
\authorrunning{V. Stankov et al.}
%
\institute{
Charles University, Faculty of Mathematics and Physics, \\ Institute of Formal and Applied Linguistics (ÚFAL),
Malostranské Náměstí 25, 11800 
Prague, Czechia
\email{\{stankov,kopp,bojar\}@ufal.mff.cuni.cz}}

\maketitle              

\renewcommand{\thefootnote}{}
\footnotetext[1]{\textbf{Data Availability:} \\
\url{http://hdl.handle.net/11234/1-5946}\\
\url{https://huggingface.co/datasets/ufal/parczech4speech-segmented}\\ 
\url{https://huggingface.co/datasets/ufal/parczech4speech-unsegmented}.}
\renewcommand{\thefootnote}{\arabic{footnote}}

\begin{abstract}
We introduce ParCzech4Speech 1.0, a processed version of the ParCzech 4.0 corpus, targeted at speech modeling tasks with the largest variant containing 2,695 hours. We combined the sound recordings of the Czech parliamentary speeches with the official transcripts. The recordings were processed with WhisperX and Wav2Vec 2.0 to extract automated audio-text alignment. Our processing pipeline improves upon the ParCzech 3.0 speech recognition version by extracting more data with higher alignment reliability. The dataset is offered in three flexible variants: (1) \textit{sentence-segmented} for automatic speech recognition and speech synthesis tasks with clean boundaries, (2) \textit{unsegmented} preserving original utterance flow across sentences, and (3) a \textit{raw-alignment} for further custom refinement for other possible tasks. All variants maintain the original metadata and are released under a permissive CC-BY license. The dataset is available in the LINDAT repository, with the sentence-segmented and unsegmented variants additionally available on Hugging Face.
\end{abstract}
\section{Introduction}
Modern speech technologies, such as automatic speech recognition (ASR) and text-to-speech synthesis (TTS), require large-scale, high-quality datasets. For instance, successful ASR systems like Whisper~\cite{whisper} and Wav2Vec 2.0~\cite{wav2vec} were trained on thousands of hours of speech. In the TTS domain, models such as VALL-E~\cite{valle} and VoiceCraft~\cite{voicecraft} also rely on extensive datasets to achieve high-quality and natural-sounding synthesis.  The English language, in particular, benefits from numerous open resources (e.g., LibriSpeech~\cite{librispeech}, LibriTTS~\cite{libritts}, TED-LIUM 3~\cite{ted-lium}, Switchboard~\cite{switchboard}, and others) that are useful for many tasks, including speech recognition, text-to-speech synthesis, unsupervised learning, and speaker identification. 

Unfortunately, other languages such as Czech lack speech datasets that would combine critical features: \textit{size}, \textit{broad topic coverage}, \textit{adaptability}, and \textit{commercial accessibility}. While Czech speech datasets exist (e.g., Czech Parliament Meetings~\cite{Czech-Parliament-Meetings}, Czech Malach Cross-lingual Speech Retrieval Test Collection~\cite{malach}, OVM – Otázky Václava Moravce~\cite{ovm}), their adoption in commercial applications is hindered by restrictive licenses (e.g., CC-NC/ND), small size (e.g., Common Voice~\cite{common-voice}, Vystadial~\cite{vystadial16}), a very specific domain (e.g., the Spoken Corpus of Karel Makoň~\cite{makon}), lack of transcriptions (e.g., VoxPopuli~\cite{voxpopuli}), or fixed alignment that cannot be adapted to new tasks (e.g., ParCzech 3.0~\cite{parczech3,kopp2021tsd}, ParlaSpeech-CZ~\cite{Kopp_ParlaSpeechCZ1.0}).

We present \textbf{ParCzech4Speech} version 1.0, a new  Czech speech dataset designed to address these gaps. Our dataset offers:
\begin{itemize}
    \item The largest (unsegmented) variant containing \textbf{2,695 hours} of automatically recognized and aligned speech from \textbf{587 speakers}.
    \item \textbf{Three task-optimized formats}: (1) sentence-segmented variant for ASR and TTS, (2) unsegmented variant for real-world streaming scenarios where speech is continuous, and (3) raw word-level alignments for custom preprocessing.
    \item \textbf{Rich metadata}, including speaker information, official and recognized transcriptions, word-level alignment, and various automatic metrics, allowing fine-grained control over pre-processing.
    \item \textbf{Commercial accessibility}: Unlike many existing resources (e.g., CC-NC licensed PDTSC~\cite{pdtsc} or Czech Malach Cross-lingual Speech Retrieval Test Collection~\cite{malach}), we release our data under CC-BY license, enabling unrestricted commercial use.
\end{itemize}

\section{Related Work}

High-quality speech datasets have traditionally relied on manual annotation, as seen in Czech resources like PDTSC 2.0~\cite{pdtsc} (122 hours of dialogues) and Vystadial~\cite{vystadial16} (15 hours of call-center speech). While these provide precise transcripts invaluable for benchmarks, their labor-intensive creation motivates the need for scalable automated alternatives.

Several approaches to automatic alignment have been used in the past. Early systems for Czech speech data employed GMM-HMM models, as seen in ParCzech 3.0 and 4.0~\cite{parczech4} for parliamentary speeches and the Spoken Corpus of Karel Makoň. These demonstrated the feasibility of automated processing, but used hybrid ASR systems, which can now be replaced by more advanced neural models.

The ParlaSpeech-CZ 1.0~\cite{Kopp_ParlaSpeechCZ1.0} dataset has been developed within the ParlaMint project~\cite{Erjavec_ParlaMint_II_advancing_2024}. The ParlaSpeech-CZ dataset allows easy use of aligned data, because it is available on Hugging Face.\footnote{https://huggingface.co/datasets/classla/ParlaSpeech-CZ} The dataset was processed using a GMM-HMM acoustic model and includes a carefully selected subset of 717,682 sentences with alignment-based truth-normalized edit distance $\le 0.15$, totaling 1,218 hours of speech. 

Recent advances have shifted toward neural approaches. Forced alignment with CTC-based~\cite{ctc} acoustic models has become standard when precise transcripts are available. The MMS-lab project~\cite{ctc-alignment-application} notably applied CTC segmentation~\cite{ctc-alignment} to create a multilingual corpus using religious texts. 

WhisperX~\cite{whisperx} represents a significant advancement by combining voice activity detection with Wav2Vec 2.0 forced alignment based on CTC segmentation. Its key innovations include audio segmentation into manageable chunks using Voice Activity Detection (VAD), followed by a Cut \& Merge strategy to create coherent audio segments, batched inference to reduce computational overhead, and precise word-level alignment. These developments address critical limitations of previous methods, mainly related to processing long audio recordings and improving alignment precision.

Building on these advances, our work applies state-of-the-art alignment techniques to Czech parliamentary data. We use modern ASR models, namely WhisperX and Wav2Vec 2.0, to create a large-scale, high-quality dataset with flexible formats for various speech modeling tasks.

\section{Dataset Construction}

Our dataset is derived from the ParCzech 4.0 corpus and the AudioPSP 24.01 collection~\cite{AudioPSP24.01}, which provides the corresponding audio recordings. ParCzech was tokenized using the UDPipe tool~\cite{straka-2018-udpipe}, which also provided sentence boundaries. We have transformed the original ParCzech files from the TEI format to the TSV formatted files that correspond to audio files in AudioPSP. One row in the TSV file corresponds to a particular token with all essential information such as token ID,\footnote{The segment-level ID compatible with ParCzech 4.0 can be used to obtain diverse pieces of information, e.g. the speaker's role in the session, their gender and date of birth.} the token text, speaker ID, date of the speech, and information on whether the token is a word or punctuation.


A known challenge with stenographic protocols is their lack of complete precision -- they may omit some spoken words as they represent official transcripts that normalize spoken language artifacts like disfluencies, repetitions, and self-corrections. This prevents using the original transcripts directly for forced alignment. To address this, we employ a two-step approach: first recognizing the audio using a robust ASR model, then performing forced alignment using the recognized transcriptions to establish audio-text alignment. The resulting timestamps allow us to map and retain only those portions of audio that match the official stenographic protocols.

We utilize the WhisperX tool for both recognition and forced alignment, employing its Wav2Vec 2.0 model for word-level timestamps. This pipeline has one limitation: while Whisper may output non-normalized text (such as numbers in digit form or special characters), Wav2Vec 2.0 cannot provide timestamps for digits and pronounced special characters (\%, \&, @, §), leaving these words without timestamp assignments.

After audio recognition and timestamp extraction, we align the recognized transcript\footnote{For recognized transcripts, we use trivial tokenization based on whitespace, so punctuation symbols typically remain attached to the neighboring word.} with the official transcript using a two-stage word-level alignment process that preserves word-level timestamps. In the first stage, the two word sequences are aligned at the word level, treating each word as an atomic unit. To determine whether a pair of words is a good match, we compute the Levenshtein distance between them. If the words differ significantly (i.e., the Levenshtein distance exceeds the length of the official word), we insert a special gap token instead of forcing the alignment of dissimilar words. This approach ensures that only sufficiently similar words are matched, while maintaining accurate timestamp propagation.

\begin{table}[tb]
    \centering
    \begin{tabular}{l|rrrr}
         & \textbf{Source} & \textbf{Recognized} & \textbf{Aligned} & \textbf{Perfectly Aligned} \\
        \hline
        Word count & 30,\,562,\,859 & 40,\,525,\,328 & 27,\,835,\,729 & 25,\,562,\,587 \\
        Word count (timed) & - & 39,\,844,\,025 & 27,\,420,\,490 & 25,\,216,\,911 \\
        Duration (hours) & 6\,431 & 4\,770 & 3\,308 & 2\,990 \\
        Unique speakers & 590 & 589 & 588 & 588 \\
    \end{tabular}
    \caption{Amounts of data obtained from the source stenographic transcripts and sound recordings before and after processing with our alignment pipeline.}
    \label{tab:alignment_stats}
\end{table}

The basic variant of our dataset provides this word-level alignment, the alignment statistics in \cref{tab:alignment_stats} quantify the processing pipeline's effectiveness. The Source variant is measured on the raw audio and text data after being processed by UDPipe.  For this variant, the word count and the number of unique speakers are constrained by the availability of corresponding audio files, meaning that statistics are computed only for data that include both transcript and audio. The Aligned variant counts only word pairs where each recognized word is matched with a word from the original transcript, i.e. neither the recognized word nor the original word is a gap word. The Perfectly Aligned column counts only those pairs where the Levenshtein distance between the recognized and original word is zero, i.e. they are exactly the same. It is important to note that the original audio files always include offset content at the beginning and end (approximately one minute on each side), which is not covered by the official transcripts. We report the word count in the Recognized variant based on full audio recordings. This explains the increased number of tokens in the Recognized variant, since the original transcript does not include the words from the offset. As was previously stated, Wav2Vec 2.0 model is not able to provide the timestamps for numbers and pronounced special characters, which results in a smaller number (not reported in the table) of recognized words that have a timestamp assigned. This discrepancy is also visible in the Aligned and Perfectly Aligned columns. After alignment, 27.8M words are successfully matched with timestamps, with 25.2M words achieving perfect alignment (Levenshtein distance is zero). Speaker coverage remains high throughout our filtering process, with only 2 speakers excluded during the processing. The duration numbers indicate that the pipeline preserves 51.3\% of original audio hours (3,308h) in the aligned variant, with 2,990 hours meeting strict quality thresholds.\footnote{The duration is computed as the sum of durations of aligned words.}

This alignment serves as the basis for the additional dataset's formats. It can also be used for custom tasks, where the user wants decide how to use the word-level timestamps. Our word-level alignment links back to the original ParCzech 4.0 and AudioPSP datasets through unique word IDs and audio file paths, stored as additional columns in the TSV files. This approach enables us to publish only the TSV files in the LINDAT repository, without needing to include the original audio or TEI files.

\subsection{Sentence-Segmented Variant}

The sentence-segmented variant is created by dividing the audio according to UDPipe sentence boundaries, using the word-level timestamps assigned during alignment. For a segment to be included, both its starting and ending words must have precise timestamps (i.e., not be aligned to gap words). While words within segments may be aligned to gap words, we conservatively discard any segment containing gap words to ensure data quality, as neither the original nor recognized transcript can be fully trusted in these cases due to potential ASR errors or manual corrections in the original transcripts. The only exception is made for special symbols and numbers, which lack timestamps due to Wav2Vec 2.0 limitations. Segments starting or ending with these non-timestamped elements are discarded. Finally, we keep only segments with a single speaker.

We employ four automatic quality metrics stored in each segment's metadata and threshold them in the filtering phase:
\begin{itemize}
    \item \textbf{Segment edit distance}: The Levenshtein distance between normalized original and recognized text segments (threshold: $\leq$9).
    \item \textbf{Maximum aligned edit distance}: The highest word-level edit distance within a segment (threshold: $\leq$5).
    \item \textbf{Average character duration}: Segment duration divided by number of characters, identifying unnatural speech rates which often indicates timestamp issues or long pauses (thresholds: 0.06s $\leq$ duration $\leq$ 0.14s).
    \item \textbf{Speaker-text counter}: Occurrences of each segment per speaker (for analysis only, not filtering).
\end{itemize}

Thresholds were determined through manual verification of random samples. This process revealed an alignment artifact: while WhisperX may produce textually perfect transcripts for number-heavy segments, the lack of timestamps for numbers can shift surrounding word timestamps, compromising boundary precision. We mitigate this by re-recognizing segments with a Wav2Vec 2.0 model and compare the Wav2Vec 2.0 and Whisper transcriptions using edit distance. Since Wav2Vec 2.0 provides only the textual transcription, for filtering purposes the numbers in the Whisper transcript were replaced with the corresponding number in the base form (e.g., ``1,000'' becomes ``tisíc''), even though this heuristic is not perfect it allows  to have a match (with larger edit distance) between the segments.

Each segment includes metadata fields for speaker ID, numeric token count, word count, and token position identifiers. The dataset preserves the test and development splits from ParCzech 3.0's speaker sets. \cref{tab:seg_stats} summarizes the statistics of the final sentence-segmented corpus, which is available on both Hugging Face and the LINDAT repository. The resulting test split misses one speaker from the original ParCzech 3.0 test set, while the development split retains all speakers.

\begin{table}[tb]
    \centering
    \begin{tabular}{l|rrr}
        & \textbf{Train} & \textbf{Dev} & \textbf{Test} \\
        \hline
        Word count & 8,\,684,912 & 74,\,556 & 164,\,803 \\
        Segments & 682,\,254 & 5 094 & 11,\,379 \\
        Duration (hours) & 1\,131 & 10.14 & 20.63 \\
        Unique speakers & 525 & 29 & 30 \\
    \end{tabular}
    \caption{Statistics of the sentence-segmented variant of ParCzech4Speech corpus.}
    \label{tab:seg_stats}    
\end{table}

We also compare our alignment procedure with the ParlaSpeech-CZ corpus. Since ParlaSpeech-CZ does not provide recognized transcripts, we simulate our cleaning procedure by applying only the average character duration filter. After filtering with our thresholds (0.06 and 0.14), the resulting ParlaSpeech-CZ dataset contains 496,405 segments with a total duration of 1040.38 hours. In contrast, applying the truth-normalized edit distance filter with a threshold of 0.15 to our segmented variant yields 689,974 segments with a total duration of 1156 hours.

This comparison shows that the two datasets are comparable in overall duration, with our additionally filtered segmented variant containing approximately 11\% more audio data than filtered ParlaSpeech-CZ corpus. Furthermore, the higher number of segments in our dataset suggests that our segments are, on average, shorter than those in ParlaSpeech-CZ.

\subsection{Unsegmented Variant}

The unsegmented variant captures continuous speech sequences without strict sentence boundaries, making it particularly suitable for real-world streaming applications where speech is unsegmented. Unlike the sentence-segmented variant, utterances are constructed by aggregating consecutive well-aligned words until encountering either a gap word or a speaker change. This approach better reflects natural speech patterns where sentences may be incomplete.

Special characters and numbers require particular attention in this variant due to their lack of timestamps. When these untimestamped elements appear at segment boundaries, we adjust to the nearest timestamped word, with a single exception for punctuation marks that immediately follow timestamped words. This ensures all retained segments have reliable start and end points while maintaining natural speech flow.

For quality control, we use the same edit distance thresholds as the sentence-segmented variant (segment edit distance $\leq$ 9, maximum aligned edit distance $\leq$ 5), but implement character duration filtering at the word level (0.035-1.0 seconds per character) rather than per-segment. We enforce minimum length requirements of 5 original words per segment, with longer segments (> 30 seconds) being randomly split while preserving this minimum word count. This balance maintains natural speech variability while preventing excessively long segments.

In this variant, we intentionally preserve realistic speech characteristics, including internal pauses and variable-length segments. We apply the same post-processing pipeline as the sentence-segmented variant, including Wav2Vec 2.0 re-recognition and edit distance filtering, to ensure consistency across both variants. Test and development splits maintain ParCzech 3.0's speaker-based partitioning, with complete statistics shown in \cref{tab:unseg_stats}. Compared to the sentence-segmented variant, the absence of sentence boundary constraints allows us to extract significantly more data from the corpus, specifically 84\% of the recognized duration (2695.8h).

\begin{table}[tb]
    \centering
    \begin{tabular}{l|rrr}
        & \textbf{Train} & \textbf{Dev} & \textbf{Test} \\
        \hline
        Word count & 18,\,385,517 & 326,\,359 & 147,\,514 \\
        Segments & 1,\,311,\,027 & 20,\,352 & 9\,127 \\
        Duration (hours) & 2\,631 & 43.43 & 21.37 \\
        Unique speakers & 527 & 30 & 30 \\
    \end{tabular}
    \caption{Statistics of the unsegmented variant of ParCzech4Speech corpus.}
    \label{tab:unseg_stats}
\end{table}

\section{Conclusions}

We presented ParCzech4Speech, a large-scale Czech speech dataset designed to address the scarcity of high-quality, freely available resources for speech modeling tasks. By leveraging modern ASR tools (WhisperX and Wav2Vec 2.0), we improved upon previous versions of the ParCzech corpus, extracting more usable data—up to 2,695 hours of speech. The dataset is released in three flexible variants: (1) sentence-segmented for ASR/TTS applications, (2) unsegmented for continuous speech scenarios, and (3) raw-alignment for custom refinements.

Our careful processing pipeline ensures high-quality alignments between the speech and transcript. We preserve metadata, including segment-level IDs to the full ParCzech 4.0 corpus and speaker information, and provide our automatic quality assessments to allow custom filtering. Unlike many existing Czech speech datasets, ParCzech4Speech is released under a CC-BY license, enabling unrestricted academic and commercial use.

Future work could explore fine-tuning ASR models on this dataset, improving alignment for numbers and special characters, or extending the corpus with additional parliamentary terms. We believe this resource will significantly benefit the Czech speech processing community and encourage further research in speech technologies.

\begin{credits}

\subsubsection{\ackname} 
The authors acknowledge the support of the National Recovery Plan funded project MPO 60273/24/21300/21000 CEDMO 2.0 NPO
and the support by the grant CZ.02.01.01/00/23\_020/0008518 (``Jazykov{\v{e}}da, um{\v{e}}l{\'{a}} inteligence a jazykov{\'{e}} a {\v{r}}e{\v{c}}ov{\'{e}} technologie: od v{\'{y}}zkumu k aplikac{\'{\i}}m'').
The work described herein has been using services provided by the LINDAT/CLARIAH-CZ Research Infrastructure (\url{https://lindat.cz}), supported by the Ministry of Education, Youth and Sports of the Czech Republic (Project No. LM2023062).

\subsubsection{\discintname}
The authors have no competing interests to declare that are relevant to the content of this article.


\end{credits}
%
%
%
\bibliographystyle{splncs04}
\bibliography{mybibliography}

\begin{thebibliography}{10}
\providecommand{\url}[1]{\texttt{#1}}
\providecommand{\urlprefix}{URL }
\providecommand{\doi}[1]{https://doi.org/#1}

\bibitem{common-voice}
Ardila, R., Branson, M., Davis, K., Kohler, M., Meyer, J., Henretty, M.,
  Morais, R., Saunders, L., Tyers, F., Weber, G.: Common voice: A
  massively-multilingual speech corpus. In: Calzolari, N., B{\'e}chet, F.,
  Blache, P., Choukri, K., Cieri, C., Declerck, T., Goggi, S., Isahara, H.,
  Maegaard, B., Mariani, J., Mazo, H., Moreno, A., Odijk, J., Piperidis, S.
  (eds.) Proceedings of the Twelfth Language Resources and Evaluation
  Conference. pp. 4218--4222. European Language Resources Association,
  Marseille, France (May 2020), \url{https://aclanthology.org/2020.lrec-1.520/}

\bibitem{wav2vec}
Baevski, A., Zhou, H., Mohamed, A., Auli, M.: wav2vec 2.0: a framework for
  self-supervised learning of speech representations. In: Proceedings of the
  34th International Conference on Neural Information Processing Systems. NIPS
  '20, Curran Associates Inc., Red Hook, NY, USA (2020)

\bibitem{whisperx}
Bain, M., Huh, J., Han, T., Zisserman, A.: Whisperx: Time-accurate speech
  transcription of long-form audio. INTERSPEECH 2023  (2023)

\bibitem{valle}
Chen, S., Wang, C., Wu, Y., Zhang, Z., Zhou, L., Liu, S., Chen, Z., Liu, Y.,
  Wang, H., Li, J., He, L., Zhao, S., Wei, F.: Neural codec language models are
  zero-shot text to speech synthesizers. IEEE Transactions on Audio, Speech and
  Language Processing  \textbf{33},  705--718 (2025).
  \doi{10.1109/TASLPRO.2025.3530270}

\bibitem{Erjavec_ParlaMint_II_advancing_2024}
Erjavec, T., Kopp, M., Ljubešić, N., Kuzman, T., Rayson, P., Osenova, P.,
  Ogrodniczuk, M., Çöltekin, {\c{C}}., Koržinek, D., Meden, K., Skubic, J.,
  Rupnik, P., Agnoloni, T., Aires, J., Barkarson, S., Bartolini, R., Bel, N.,
  Calzada~Pérez, M., Darģis, R., Diwersy, S., Gavriilidou, M., van Heusden,
  R., Iruskieta, M., Kahusk, N., Kryvenko, A., Ligeti-Nagy, N., Magariños, C.,
  Mölder, M., Navarretta, C., Simov, K., Tungland, L.M., Tuominen, J., Vidler,
  J., Vladu, A.I., Wissik, T., Yrjänäinen, V., Fišer, D.: {ParlaMint II:
  advancing comparable parliamentary corpora across Europe}. Language Resources
  and Evaluation  (Dec 2024). \doi{10.1007/s10579-024-09798-w}

\bibitem{malach}
Galu{\v s}{\v c}{\'a}kov{\'a}, P., Pecina, P., Hoffmannov{\'a}, P., Haji{\v c},
  J., Ircing, P., {\v S}vec, J.: Czech malach cross-lingual speech retrieval
  test collection (2017), \url{http://hdl.handle.net/11234/1-1912},
  {LINDAT}/{CLARIAH}-{CZ} digital library at the Institute of Formal and
  Applied Linguistics ({{\'U}FAL}), Faculty of Mathematics and Physics, Charles
  University

\bibitem{switchboard}
Godfrey, J.J., Holliman, E.C., McDaniel, J.: Switchboard: telephone speech
  corpus for research and development. In: Proceedings of the 1992 IEEE
  International Conference on Acoustics, Speech and Signal Processing - Volume
  1. p. 517–520. ICASSP'92, IEEE Computer Society, USA (1992)

\bibitem{ctc}
Graves, A., Fern\'{a}ndez, S., Gomez, F., Schmidhuber, J.: Connectionist
  temporal classification: labelling unsegmented sequence data with recurrent
  neural networks. In: Proceedings of the 23rd International Conference on
  Machine Learning. p. 369–376. ICML '06, Association for Computing
  Machinery, New York, NY, USA (2006). \doi{10.1145/1143844.1143891},
  \url{https://doi.org/10.1145/1143844.1143891}

\bibitem{AudioPSP24.01}
Kopp, M.: {AudioPSP} 24.01: Audio recordings of proceedings of the chamber of
  deputies of the parliament of the czech republic (2024),
  \url{http://hdl.handle.net/11234/1-5404}, {LINDAT}/{CLARIAH}-{CZ} digital
  library at the Institute of Formal and Applied Linguistics ({{\'U}FAL}),
  Faculty of Mathematics and Physics, Charles University

\bibitem{parczech4}
Kopp, M.: {ParCzech} 4.0 (2024), \url{http://hdl.handle.net/11234/1-5360},
  {LINDAT}/{CLARIAH}-{CZ} digital library at the Institute of Formal and
  Applied Linguistics ({{\'U}FAL}), Faculty of Mathematics and Physics, Charles
  University

\bibitem{Kopp_ParlaSpeechCZ1.0}
Kopp, M., Ljube{\v s}i{\'c}, N.: Parliamentary spoken corpus of {Czech}
  {ParlaSpeech}-{CZ} 1.0 (2024), \url{http://hdl.handle.net/11356/1785}

\bibitem{parczech3}
Kopp, M., Stankov, V., Bojar, O., Hladk{\'a}, B., Stra{\v n}{\'a}k, P.:
  {ParCzech} 3.0 (2021), \url{http://hdl.handle.net/11234/1-3631},
  {LINDAT}/{CLARIAH}-{CZ} digital library at the Institute of Formal and
  Applied Linguistics ({{\'U}FAL}), Faculty of Mathematics and Physics, Charles
  University

\bibitem{kopp2021tsd}
Kopp, M., Stankov, V., Kr{\r{u}}za, J., Stra{\v{n}}{\'{a}}k, P., Bojar, O.:
  {ParCzech} 3.0: A large {Czech} speech corpus with rich metadata. In:
  Ek{\v{s}}tein, K., P{\'{a}}rtl, F., Konop{\'{\i}}k, M. (eds.) Text, Speech,
  and Dialogue. pp. 293--304. Lecture Notes in Computer Science, University of
  West Bohemia, Springer, Cham, Switzerland (2021).
  \doi{10.1007/978-3-030-83527-9_25}

\bibitem{makon}
Kr{\r u}za, J.O.: Spoken corpus of karel mako{\v n} (2020-11-16) (2020),
  \url{http://hdl.handle.net/11234/1-3422}, {LINDAT}/{CLARIAH}-{CZ} digital
  library at the Institute of Formal and Applied Linguistics ({{\'U}FAL}),
  Faculty of Mathematics and Physics, Charles University

\bibitem{ctc-alignment}
K{\"u}rzinger, L., Winkelbauer, D., Li, L., Watzel, T., Rigoll, G.:
  Ctc-segmentation of large corpora for german end-to-end speech recognition.
  In: Karpov, A., Potapova, R. (eds.) Speech and Computer. pp. 267--278.
  Springer International Publishing, Cham (2020)

\bibitem{pdtsc}
Mikulov{\'a}, M., B{\'e}mov{\'a}, A., Haji{\v c}, J., Haji{\v c}ov{\'a}, E.,
  Ircing, P., Kol{\'a}{\v r}ov{\'a}, V., Lopatkov{\'a}, M., Mare{\v c}ek, D.,
  M{\'{\i}}rovsk{\'y}, J., Nedoluzhko, A., Pajas, P., Panevov{\'a}, J.,
  Peterek, N., Romportl, J., Sgall, P., {\v S}ev{\v c}{\'{\i}}kov{\'a}, M., {\v
  S}t{\v e}p{\'a}nek, J., Ure{\v s}ov{\'a}, Z., {\v Z}abokrtsk{\'y}, Z.: Prague
  dependency treebank of spoken czech 2.0 ({PDTSC} 2.0) (2017),
  \url{http://hdl.handle.net/11234/1-3189}, {LINDAT}/{CLARIAH}-{CZ} digital
  library at the Institute of Formal and Applied Linguistics ({{\'U}FAL}),
  Faculty of Mathematics and Physics, Charles University

\bibitem{librispeech}
Panayotov, V., Chen, G., Povey, D., Khudanpur, S.: Librispeech: An asr corpus
  based on public domain audio books. In: 2015 IEEE International Conference on
  Acoustics, Speech and Signal Processing (ICASSP). pp. 5206--5210 (2015).
  \doi{10.1109/ICASSP.2015.7178964}

\bibitem{voicecraft}
Peng, P., Huang, P.Y., Li, S.W., Mohamed, A., Harwath, D.: {V}oice{C}raft:
  Zero-shot speech editing and text-to-speech in the wild. In: Ku, L.W.,
  Martins, A., Srikumar, V. (eds.) Proceedings of the 62nd Annual Meeting of
  the Association for Computational Linguistics (Volume 1: Long Papers). pp.
  12442--12462. Association for Computational Linguistics, Bangkok, Thailand
  (Aug 2024). \doi{10.18653/v1/2024.acl-long.673},
  \url{https://aclanthology.org/2024.acl-long.673/}

\bibitem{vystadial16}
Pl{\'a}tek, O., Du{\v s}ek, O., Jur{\v c}{\'{\i}}{\v c}ek, F.: Vystadial 2016
  – czech data (2016), \url{http://hdl.handle.net/11234/1-1740},
  {LINDAT}/{CLARIAH}-{CZ} digital library at the Institute of Formal and
  Applied Linguistics ({{\'U}FAL}), Faculty of Mathematics and Physics, Charles
  University

\bibitem{ctc-alignment-application}
Pratap, V., Tjandra, A., Shi, B., Tomasello, P., Babu, A., Kundu, S., Elkahky,
  A., Ni, Z., Vyas, A., Fazel-Zarandi, M., Baevski, A., Adi, Y., Zhang, X.,
  Hsu, W.N., Conneau, A., Auli, M.: Scaling speech technology to 1,000+
  languages. J. Mach. Learn. Res.  \textbf{25}(1) (Jan 2024)

\bibitem{Czech-Parliament-Meetings}
Pra{\v z}{\'a}k, A., {\v S}m{\'{\i}}dl, L.: Czech parliament meetings (2012),
  \url{http://hdl.handle.net/11858/00-097C-0000-0005-CF9C-4},
  {LINDAT}/{CLARIAH}-{CZ} digital library at the Institute of Formal and
  Applied Linguistics ({{\'U}FAL}), Faculty of Mathematics and Physics, Charles
  University

\bibitem{whisper}
Radford, A., Kim, J.W., Xu, T., Brockman, G., McLeavey, C., Sutskever, I.:
  Robust speech recognition via large-scale weak supervision. In: Proceedings
  of the 40th International Conference on Machine Learning. ICML'23, JMLR.org
  (2023)

\bibitem{ted-lium}
Rousseau, A., Del{\'e}glise, P., Est{\`e}ve, Y.: {TED}-{LIUM}: an automatic
  speech recognition dedicated corpus. In: Calzolari, N., Choukri, K.,
  Declerck, T., Do{\u{g}}an, M.U., Maegaard, B., Mariani, J., Moreno, A.,
  Odijk, J., Piperidis, S. (eds.) Proceedings of the Eighth International
  Conference on Language Resources and Evaluation ({LREC}'12). pp. 125--129.
  European Language Resources Association (ELRA), Istanbul, Turkey (May 2012),
  \url{https://aclanthology.org/L12-1405/}

\bibitem{ovm}
{\v S}m{\'{\i}}dl, L., Pra{\v z}{\'a}k, A.: {OVM} – ot{\'a}zky v{\'a}clava
  moravce (2013), \url{http://hdl.handle.net/11858/00-097C-0000-000D-EC98-3},
  {LINDAT}/{CLARIAH}-{CZ} digital library at the Institute of Formal and
  Applied Linguistics ({{\'U}FAL}), Faculty of Mathematics and Physics, Charles
  University

\bibitem{straka-2018-udpipe}
Straka, M.: {UDP}ipe 2.0 {P}rototype at {C}o{NLL} 2018 {UD} {S}hared {T}ask.
  In: Proceedings of the {C}o{NLL} 2018 ST: Multilingual Parsing from Raw Text
  to Universal Dependencies. pp. 197--207. Association for Computational
  Linguistics (2018), \url{http://dx.doi.org/10.18653/v1/K18-2020}

\bibitem{voxpopuli}
Wang, C., Riviere, M., Lee, A., Wu, A., Talnikar, C., Haziza, D., Williamson,
  M., Pino, J., Dupoux, E.: {V}ox{P}opuli: A large-scale multilingual speech
  corpus for representation learning, semi-supervised learning and
  interpretation. In: Zong, C., Xia, F., Li, W., Navigli, R. (eds.) Proceedings
  of the 59th Annual Meeting of the Association for Computational Linguistics
  and the 11th International Joint Conference on Natural Language Processing
  (Volume 1: Long Papers). pp. 993--1003. Association for Computational
  Linguistics, Online (Aug 2021). \doi{10.18653/v1/2021.acl-long.80},
  \url{https://aclanthology.org/2021.acl-long.80/}

\bibitem{libritts}
Zen, H., Dang, V., Clark, R., Zhang, Y., Weiss, R.J., Jia, Y., Chen, Z., Wu,
  Y.: Libritts: A corpus derived from librispeech for text-to-speech (2019),
  \url{https://arxiv.org/abs/1904.02882}

\end{thebibliography}

\end{document}